# A New Approach to Draw Detection by Move Repetition in Computer Chess Programming

By **Vladan Vuckovic** [1] and **Djordje Vidanovic** [2],
*University of Nis, Serbia & Montenegro*

*Abstract:* We will try to tackle both the theoretical and practical aspects of a very important problem in chess programming as stated in the title of this article – the issue of draw detection by move repetition. The standard approach that has so far been employed in most chess programs is based on utilising positional matrices in original and compressed format as well as on the implementation of the so-called bitboard format.

The new approach that we will be trying to introduce is based on using variant strings generated by the search algorithm (searcher) during the tree expansion in decision making. We hope to prove that this approach is more efficient than the standard treatment of the issue, especially in positions with few pieces (endgames). To illustrate what we have in mind a machine language routine that implements our theoretical assumptions is attached. The routine is part of the *Axon* chess program, developed by the authors. *Axon*, in its current incarnation, plays chess at master strength (ca. 2400-2450 Elo, based on both *Axon* vs computer programs and *Axon* vs human masters in over 3000 games altogether).

## 1. INTRODUCTION

The solution to the problem of move repetition detection is very important for the creation of a computer chess playing program that follows and implements all official rules of chess in its searcher **[1],[2]**. Namely, in accordance with the official rules of chess there are two major types of positions in which a draw is proclaimed: (a) when the position in which the same side to move has been repeated three times in the course of a game and (b) when no capture has been made or no pawns have been moved during the last fifty consecutive moves. Having in mind that condition (b) can be implemented comparatively easily by using the data from the move generator, we will focus more closely on condition (a).

The essence of the draw detection rule strives to prevent "endless" games and to offer a possibility for a game to progress by means of non-reversible moves such as pawn movement or piece exchange and captures. Naturally, in a game of average length a relatively high percentage of moves is due to maneuverability of piece moves which lead to progress in games through a genuine transformation.

---

[1] Main author and developer of the Axon chess engine; Faculty of Electrical Engineering, University of Nis, Serbia & Montenegro. *E-mail:* **vladan@bankerinter.net**
[2] Co-author, joined the Axon team in 2003; Faculty of Philosophy, University of Nis, Serbia & Montenegro. *E-mail*: **djordjev@medianis.net**

Bearing in mind the nature of the game of chess it is not merely sufficient to investigate only positions stemming in the course of the tree search generation (decision making) **[6]**, but also to consider the game set-up prior to the actual position - the so-called "game history".

Thus we intend to offer a theoretical and practical solution to the issues we mentioned. There will be several sections dealing with various aspects of draw detection. After this initial exposition, our next section will describe the classical approach to draw detection based on the comparison of matrices. The third section will present a new treatment that employs variant strings. In the fourth section we plan to lay out a blueprint for a solving algorithm. The fifth part of the article will include a description of a routine written in assembly code implementing the discussed algorithm, while the concluding section will deal with some experimental data relevant to the algorithm as implemented in the *Axon* chess program. We will compare the behaviour of the program with and without the implemented procedure for draw detection.

## 2. STANDARD APPROACH TO THE SOLUTION OF THE PROBLEM

In the process of recognition of draw detection and recurrently identical positions chess programs employ the same memory structure used as an input for the move generator. This procedure aims at generating all legal moves stemming from the current position in any part of the search tree. Regardless of the type of piece coding which can vary a lot its common property is the existence of a positional matrix that memorizes the current position found in a part of the search tree. Besides such uncompressed forms of position representation there are various models using compression as well as models which represent positions in readable format, e.g. epd and FEN representations. Even though they require a lesser amount of memory these formats are practically unusable, especially in endgame positions where one finds many empty squares.

What happens here is that the time needed for compression and/or decompression into/out of such formats is significantly longer than the time spent in mere comparison, which leads us to think that they are not an efficient solution, particularly with regard to the speed of execution of the procedures being processed. On the other hand, we need to solve the problem of the internal representation of the chess board and piece coding in an efficient way too. The representational organization of the chess board **[1]**,**[3]** depending on the move generation procedure, can be done in different ways, for example as an 8X8, 10X10, 12X10 or 12X12 byte grid. Our program, *Axon*, has a 12X12 byte matrix structure **[7]**. We should mention though that it is only to be expected that performance-wise the best results are obtained with an 8X8 grid that is very much like an iconic representation of the chess board.

Implementations of other representational matrices lead to proportional slowdown, but the gains earned by the use of larger grids as compared to the 8X8 grid are faster recognition of the grid borders, one dimensional coding in place of two dimensional one, etc.

In order to explain the core problem better, let us propose a simple type of coding (8X8 bytes). (Let us emphasize first that the manner of piece coding is completely irrelevant to our further analysis and serves only as an illustration.) The chess board hosts 12 different pieces (six white and six black pieces), therefore to represent all of them we need four bit coding. The coding can be done in the following way:

Table 1.  *Piece coding.*

| PIECE | CODE | Dec. |
| --- | --- | --- |
| Empty square | oooo | o |
| White pawn | ooo1 | 1 |
| White knight | oo1o | 2 |
| White bishop | oo11 | 3 |
| White rook | o1oo | 4 |
| White queen | o1o1 | 5 |
| White king | o11o | 6 |
| Black pawn | 1oo1 | 9 |
| Black knight | 1o1o | 1o |
| Black bishop | 1o11 | 11 |
| Black rook | 11oo | 12 |
| Black queen | 11o1 | 13 |
| Black king | 111o | 14 |

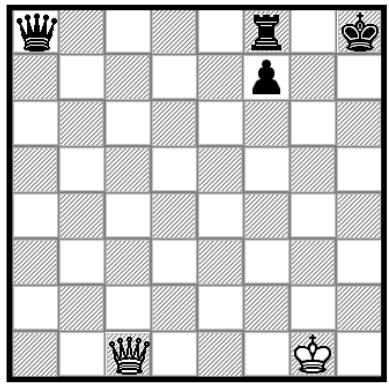

Figure 1.  White draws by perpetual check.

Figure 1 can serve as an example of a simple position in which White draws by a perpetual check (1. Qh6+). If we use the coding presented in Table 1 the internal representation of the 8X8 matrix would look like this:

| 13 | o | o | o | o | 12 | o | 14 |
|----|---|---|---|---|----|---|----|
| o  | o | o | o | o | 9  | o | o  |
| o  | o | o | o | o | o  | o | o  |
| o  | o | o | o | o | o  | o | o  |
| o  | o | o | o | o | o  | o | o  |
| o  | o | o | o | o | o  | o | o  |
| o  | o | o | o | o | o  | o | o  |
| o  | o | 5 | o | o | o  | 6 | o  |

Figure 2: Coded form of a position (Figure 1).

The positional matrix in its uncompressed format shown in Figure 2 can be used directly as input to an efficient move generator. This matrix is simultaneously part of the position detection procedure. The capacity of the matrix is 64X8-bit, equaling 512 bits, and that position comparison has to be done in the same format making the procedure inefficient. The downside to this approach can be seen rather clearly in the endgame where great portions of the chessboard are empty squares and a great deal of must be spent on comparisons of relatively similar positions. Another disadvantage is that in long lasting games, with many moves played in the endgame, one has to take into account the length of the "game history" so that the analysis may be extended to tens of other, already played, positions. We will list some of the methods that can help to speed up or eliminate the inflated 512-bit comparison:

- In the search stage we start with a position that is currently being evaluated as the deepest in the search tree. The investigation extends towards the root of the search tree. If there is some sort of "relevant history" the search must be extended onto positions stored in the corresponding data structures. If a repeated position is found (such cases are rare from a statistical point of view) the moves that might otherwise be produced by this node are not generated and the evaluation is not assigned. The value of 0.00 is automatically put down as the evaluation mark instead. Since the comparison is done only in the case when the same side is to move it is necessary to take into account only positions with either even or odd depths depending on whether the current position that serves as a start point is of even or odd ply depth. This helps to reduce the number of possible search queries by 50%.

- The search has to be terminated at the moment a position that resulted from an exchange of pieces or pawn advancement has been registered. If no position repetition had been registered up to that moment, a conclusion has to be drawn that there was no position repetition whatsoever.

- In order to speed up the search a positional checksum (16-bit) can be introduced so that the control checksum should be verified before any comparison of positions. In case the sums do not check out correctly there is no need to investigate complete matrices: they simply have to differ. As checksum a part of the hash key variable can be used, of course if the program has viable hash support.

- Using the 32-bit architecture of modern processors 32-bit registers can also be exploited for comparison purposes. In this manner, we could ensure 16 comparisons at least. With the latest 64 bit processors, we could have 8 comparisons at the most. Machine code can contribute significantly to such speed ups.

Most of the recently available chess engines have procedures developed on the basis of principles sketched in this short review. It appears to us that it is quite possible to introduce a more efficient algorithm which would avoid the inflated 512 bit structure, by relying on variant strings of moves coded in the 16 bit format.

### 3. NEW APPROACH BASED ON VARIANT STRINGS

During the tree expansion when any regular search algorithm that follows the rule of regular side to move change is employed (such as *AlphaBeta* **[1],[4],[8]** *PVS*, *NegaScout*; null move **[5]** is excluded here as it permits one side to execute two moves in a turn) so-called variant strings are generated. These strings are actually move lists that start at the root of the search tree and end up in the terminal node being evaluated. This concept is quite similar to the concept of best-line. As a matter of fact, best-line is a variant that contains best moves for both sides (own side and the opponent). Depending on the manner of coding there might be different types of variant strings. For instance, the main draw string in Figure 1 can be presented as follows:

Qh6+ Kg8 Qg5+ Kh7 Qh5+ Kg7 Qg5+ Kh8 Qh6+ …

Alternatively, we could use a similar, but basically equivalent, positional notation applied in many chess programs (most notably the Winboard GUI) (striving to unify and compress information):

C1H6 H8G8 H6G5 G8H7 G5H5 H7G7 H5G5 G7H8 G5H6 …

Using positional notation it is easy to obtain information regarding the from- and to- squares (starting and destination piece movements). If we take only a brief glance we can see that 4 bytes (32 bits) are necessary to code one single move. However, it is possible to reduce the number of bytes to 16 in coding one move. Viewing a positional matrix that has 8X8 squares, 6 bits (or 64 combinations) are necessary to code these squares. In case of a 12X12 matrix 144 squares or 8 bits are required.

This means that a 16-bit system of coding can be defined in a way that the from- and to- squares are coded by using 8 bits.

Such a simple coding system has served as a basis for an original procedure for draw detection by help of variant strings. The basic premise is that the information helping draw detection is not obtained via positions but by help of variants coded in the previous section. In *Axon*, our chess playing program, the organization of the move-generator and the tree search system is embedded in the type of coding we have just described in general. Variant strings that are formed in the tree search all the way to the terminal nodes serve as input to the move repetition procedure. The next section will describe and analyze the algorithm we used to achieve this goal and thus speed up the program itself.

## 4. ALGORITHM

A great majority of contemporary computer chess searchers has evolved from some sort of recursion. The basic types of algorithms that have already been mentioned have also been based on recursive procedures. *Axon* has a recursive kernel which makes the recognition of repetition detection a little more complex as variant strings are formed within the main program stack together with other important parameters crucial to the searcher (Alpha Beta values, positions, return value addresses, etc.). Since the repetition detection needs access to the variant string starting from the terminal node backwards, this information is not easy to reach through the stack, which is commonplace in recursive programs.

On the other hand, it is quite handy to use the same procedure together with the game history (a list of the previously played moves) so as to maximise the efficiency of the procedure. Having in mind such issues, the authors are willing to suggest the following solution (the string annotation is in the standard programming language Pascal):

The ancillary string is introduced:

list_of_moves : *array [0..254] of word;*

The string *list_of_moves* is supposed to memorise the game history as well as the current string generated as a result of the tree calculation. As a matter of fact, the mentioned string contains 16-bit coded moves starting with the initial move up to the terminal node. Before the initialisation of the iterative searcher the procedure *Generate_list_of moves* is invoked and it resets the whole string reading into it 16-bit coded moves taken over from the ancillary string containing the game history. Such a string is being updated during the activity of the searcher so that the procedure enabling the move generator on the basis of the search depth writes the coded moves directly into the string marked as *list_of_moves*. The address (*offset*) of the element that was put down last is written into the variable *move_TOS* of the *word* type. Defined in this manner, the data represent input parameters that generate the procedure dubbed *PERPETUAL* whose details will be presented in the next section.

*The move coding* generated in the procedure *move_generator* is done on the basis of an inherently spatial principle. To speed up the coding, *Axon* employs a spatial one-dimensional 8-bit address of the from- and to- square. To illustrate this, let us return to the previous section and the first move in the draw line C1H6 (Figure 1). If we code the chessboard unidimensionally so that the A8 square has a value of 0, the B8 square value 1 and the H1 square value of 63 then the move C1-H6 can be coded as 58-23, i.e. instead of two 8-bit numbers it can be replaced by one 16-bit number whose value is 14871 (58*256+23). All other moves can be coded in the same way. If *move_generator* has generated any irreversible move such as a capture, pawn advance or castling this can be marked by setting the bit of the greatest possible weight in the 8-bit code of the from- square. Such an action is possible because we need only 6 bits ($2^6$=64) to code the square itself.

The main procedure *PERPETUAL* aims to process the structure marked as *list_of_moves* and on the basis of that to return one of the two possible logical values: the value of 0 if move repetition has been detected and the value that is different from 0 if not. This procedure hinges on the utilisation of the concatenating list:

chain_list : *array [0..24] of word;*

The algorithm of the *PERPETUAL* procedure can be described using the following steps:

1) The initial indicator is set to the value *move_TOS* - the address of the last move written in the move list;

2) All the elements of the concatenating list *(chain_list)* are set to 0;

3) Scanning the move list starts from the last address *(move_TOS)* towards the lower addresses. In case the list yields an element whose bit is of the greatest weight marked 1 the procedure is abandoned immediately as it is an irreversible move meaning that position repetition has not been detected.

4) If the move was reversible it is included in the concatenating list, observing the following rules:

- If the concatenating list has a move whose to- square is identical with the from- square of the current move then the concatenation is performed by way of the transformation of the moves in the concatenating list. An example: if the concatenating list contains the move A1B1 and the current move is B1E5 then instead of the existing A1B1 the move A1E5 is listed. This rule can be dubbed *the rule of* transition as it is equivalent to the corresponding relation in mathematics. If, after the application of the rule of transition, a complete correspondence between the from- and to- square has been noted the move is removed from the concatenating list (*the rule of* reflexion). An example: if the move A1B1 is in the concatenating list and the current move is B1A1 the application of transition should generate A1A1 so that instead of that the corresponding slot in the concatenating list is filled up by a zero. The basic logic of the procedure is *closing the variant chains. Only in case that all the chains are closed (in other words all elements of the concatenating list are equal to 0) can move repetition be recognised.* Let us illustrate this point by going back to Figure 1. Suppose that the following line is possible: C1H6 H8G8 H6G5 G8H8 G5H6 … Since the scanning is done backwards it is easy to see that after the first two steps two potential chains of moves are possible (G5H6 and G8H8), after the third step the first chain is closed (H6G5), while the fourth step marks the closing of the second chain (H8G8), so that a we can proclaim a case of move repetition. Thus if all variant chains in the concatenating list close at any point then we have move repetition. This procedure is repeated after every move from the move list. If the bottom of the list or any move that has a bit set to the greatest weight is reached the procedure is terminated as move repetition has not been detected. The procedure appears quite efficient and encompasses even far more complicated positions that exhibit multivariant chains. The following figure presents a chess study that can be our testing ground where the efficiency of our move repetition detection method can be assessed:

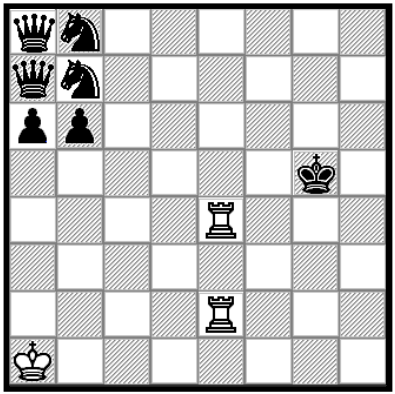

Figure 2.

It is quite clear that Black has enough potential to free its pieces and to win in just one move. If White is to move it can draw by simply checking the black king, using the rooks along the e-file.  It is curious to note that the program needs some time and a very deep calculation to realise that, regardless of the complex routes that the black king can take along the three free verticals, the position should be evaluated as 0.00. Such positions are very good at showing how advantageous the new principle of move repetition detection is over the traditional and standard one.  The new principle is based on the highly efficient 16-bit variant chains that could contain dozens of elements but are still less costly than  a great number of very similar 512-bit positions.

## 5.  IMPLEMENTATION

This section will start with a commented listing of a machine code portion that implements the algorithm that we discussed in the previous section. The procedure is part of the chess program *Axon* whose kernel has been completely coded in assembly. The procedure is very simple and, as opposed to the approach discussed in section 2, where 512-bit data structure comparisons were described, employs only 16-bit comparisons.  Theoretically and practically viewed, speedups in the used procedure when compared with the matrix approach while solving one and the same problem are considerable, especially in the endgame. The parameters of the procedure are as follows:

**Input:**

manevar_TOS: word;              {offset addresses of the top of the variant stack}
list_of_moves: array of word;   {list of moves}

**Output:**

        CX: registar;              {contains value 0 if move repetition has been found or a value different from 0 if that is not the case}

**procedure PERPETUAL;assembly;**
**asm**

```
mov si,[move_TOS]            {A pointer is read into the SI register onto the top of
                              the variant stack}
    mov di,offset chain_list {A pointer is read into the DI register into the
                              concatenating list}
    xor ax,ax                {AX=0, AX register stands for the counter of active
                              elements in the concatenating list}
    mov cx,24                {CX=24, dimension of the concatenating list}
    rep stosw                {The concatenating list is filled with zeroes}

@main:                       {Beginning of the main loop}

    mov dx,[si]              {The coded move from the move list is read into DX}
    test dh,80h              {If it is a pawn move or an exchange no draw detected}
    jnz @no_draw
    sub si,2                 {Moves back to the preceding stack element}

@mainscan:

    mov di,offset chain_list {DI is a pointer at the beginning of the concatenating list}
    mov cx,24                {CX marks the capacity of the list}

@ms1:                        {Loop that scans the concatenating list}
    cmp dl,[di+1]            {Is the to- square from DX the from- square of a move
                              in the concatenating list?}
    jz @chain
    add di,2                 {Increase of the pointer in the concatenating list}
    loop @ms1                {Control of the end of the loop}

@mainzero:                   {In case that a corresponding move has not been found
                              in the concatenating list, the list is scanned}
```

```
    xor bx,bx                      {BX=0}
    mov di,offset chain_list       {DI  is the pointer at the beginning of the concatenating list}
    mov cx,24                      {CX  is the capacity of the list}
@ms2:
    cmp [di],bx                    {Search for the free element of the concatenating list}
    jz @add_node                   {If this element is found, a new one is added to the list}
    add di,2                       {Next element…}
    loop @ms2                      {The counter of elements}
    jmp @no_draw                   {If there are not any other free elements in the list, the exit is @no_draw}

@draw:      xor cx,cx              {Draw is found, output register CX is zero}
            jmp @retq
@no_draw:   mov cx,1               {Draw is not found, output register CX = 1}
@retq:      RET                    {Procedure exit}

@add_node:                         {Addition of a new element to the concatenating list}

    mov [di],dx                    {New element is being set up…}
    inc ax                         {The counter of active elements is increased by 1}
    jmp @main                      {Jump to the beginning of the main loop}

@chain:

    cmp dh,[di]                    {If a closed cycle of moves is an issue, that element is reset in the list – filled by a zero}

    jz @no_chain

@re_chain:

    mov [di+1],dh                  {In other cases the list is being concatenated}
    jmp @main

@no_chain:                         {If deletion of elements is an issue…}

    dec ax                         {Decrease of the counter of active elements by 1}
    jz @draw                       {If there are no other active elements that means that all combinatorial chains have been closed and that a case of move repetition has been detected – Draw}
 mov word ptr [di],0
 jmp @main
```

**end;**

The procedure we have just described is very efficient in its execution and appears to be superior to the already existing modalities of move repetition and draw detection due to the advanced algorithm that was implemented in *Axon.*

In practical play the significance of this procedure is even greater especially in endgames with rooks or queens and exposed kings where checking sequences can be very long. Significant portions of the search tree in which move repetition is processed are not considered but are simply replaced by the zero value. In this manner the tree search speed is enhanced indirectly. [3]

## 5. EXPERIMENTAL DATA

In order to illustrate the influence of the described procedure on the execution of the *Axon* chess program we will use a chess position from the WAC.epd chess suite that is widely spread among computer chess programmers. It is a rook ending with both kings exposed (Figure 3). Similar positions appear rather frequently in over-the-board tournament chess. The version of *Axon* we used was a slightly modified one with a null move generator implemented as well as some other pruning techniques. On the other hand we turned off the heuristic knowledge of endgames so as to make the results of the search dependent almost solely on the search algorithm **[9]**. Endgame tablebases (such as the Nalimov tablebases) were not used. The hardware platform was a PC running on an AMD Athlon 2200+ processor with 256 Mb RAM. The test was actually an analysis of the position with the algorithm for move repetition turned on and off. The search depth was iteratively increased from ply 1 to ply 12 in both modalities (move repetition algorithm turned on and off) and the recorded parameters were: best key move at a given depth, evaluation, node count, total count of terminal and generated positions in both modalities, as well as the relative speed of generation and the length of the time spent in search.

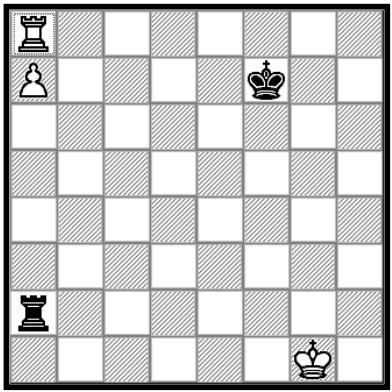

Figure 3. White to move.

The solution leading to the winning sequence is, naturally, Rh8!, thus making it possible for the white king to make a smooth journey to the solitary white pawn which will soon be promoted to a queen. Meantime the black rook can check the exposed

---
[3] See Appendix for an illustrative game that Axon played against a program estimated to be >2650 ELO at the sudden death level of game in 10 minutes (Diep-Axon).

white king for quite some time. The following tables present the data obtained from the program:

**Table 2.** *The analysis of the position **without** the move repetition algorithm on*

| Depth (ply) | Key move | Evaluation | Node count |
|---|---|---|---|
| 1 | G1F1 | +2.30 | 929 |
| 2 | G1F1 | +2.23 | 1396 |
| 3 | G1F1 | +2.44 | 7299 |
| 4 | G1F1 | +2.37 | 18964 |
| 5 | G1F1 | +2.42 | 43701 |
| 6 | A8H8 | +2.47 | 274992 |
| 7 | G1F1 | +2.48 | 364926 |
| 8 | G1F1 | +2.52 | 1541540 |
| 9 | A8H8 | +2.55 | 2252388 |
| 10 | A8H8 | +2.55 | 2062293 |
| 11 | A8H8 | +2.65 | 4635873 |
| 12 | A8H8 | +5.04 | 9770293 |

The total count was: 1 193 867 terminal nodes evaluated and 20 974 594 positions generated.

**Table 3.** *The analysis of the position **with** the move repetition algorithm turned on*

| Depth (ply) | Key move | Evaluation | Node count |
|---|---|---|---|
| 1 | G1F1 | +2.30 | 929 |
| 2 | G1F1 | +2.23 | 1396 |
| 3 | G1F1 | +2.44 | 7299 |
| 4 | G1F1 | +2.37 | 18883 |
| 5 | G1F1 | +2.41 | 46990 |
| 6 | A8H8 | +2.47 | 263862 |
| 7 | G1F1 | +2.48 | 399843 |
| 8 | A8H8 | +2.47 | 2806126 |
| 9 | A8H8 | +2.55 | 693285 |
| 10 | A8H8 | +2.55 | 1451126 |
| 11 | A8H8 | +2.55 | 3618668 |
| 12 | A8H8 | +4.87 | 8524217 |

In this case the total count was: 977942 terminal nodes evaluated, 17 832 624 positions generated. 14 941 cases of position repetitions were noted.

The conclusions to be drawn from the data are as follows:

The structure of the key moves and evaluation in both experiments was similar, the differences are generated because of the activity of the extending search algorithm in *Axon* that adapts to the search depth (adaptive depth control).

- The comparative increase of the positions searched is similar in both cases. However, the increase of search depth the move repetition algorithm is evidently beneficial here. This is so because the repeated position is detected at a shallower depth and such positions are evaluated as 0.00 so that the sub-trees stemming from them are pruned and not extended any further as is the case in the first part of the experiment. The second part of the experiment shows that the algorithm is slower by about 10%, the time taken up by the move repetition procedure. The decrease in speed is smaller in the middle game or in positions where there are many non-reversible moves.
- The tree being generated in the second part of the experiment is smaller by 18% judging by the number of terminal positions, and by 15% counting the number of generated positions. The analysis performed in the second modality has a shorter duration by 10% than the one in the first modality with the move repetition algorithm off.

The conclusion to be drawn from our experiment is that the move repetition detection procedure that is based on the method utilised by the authors processes fewer positions and creates a smaller number of terminal nodes thus compensating for the loss of general processing speed. The procedure is beneficial in terms of overall performance achieving a gain of approximately 10%. In endgames with queens and exposed kings this percentage should prove greater. The authors hope to have presented credible and convincing experimental data to justify the inclusion of their move repetition detection algorithm into state-of-the-art computer chess programs.

## 6. CONCLUSION

The novelty that the authors have proposed in this paper relates to a new approach to the problem of move repetition detection that is based on variant strings as opposed to the matrix-based approach used in the majority of the existing chess playing programs. The authors have attempted to prove the efficiency of their treatment of the problem in both the areas of implementation and the speed of execution. The algorithm proposed by the authors has been discussed in detail, with portions of their chess playing program *Axon* presented in machine code. *Axon* was also used in a small experiment relating to the discussed issue of move repetition detection by means of which the authors contend that the implementation of their procedure leads to considerable gain (10% overall) in the processing of the search tree, especially in endgames. As the program used in the experiment (*Axon*) does not have the positional (matrix-based) algorithm for move repetition detection, it was not possible for the authors to directly compare the two methods, but they firmly believe that the segment of analysis presented may convince the reader about its advantages.

## 8. APPENDIX

[White "DIEP version=2.00"]
[Black "Axon"]
[Result "1/2-1/2"]

1. d4 Nf6 2. Bf4 d5 3. Nd2 Nc6 4. e3 Bf5 5. Bb5 a6 6. Bxc6+ bxc6 7. Ne2 Rb8 8. b3 e6 9. O-O Bb4 10. Nf3 O-O 11. Ne5 Rb6 12. c4 dxc4 13. bxc4 Bd6 14. c5 Bxe5 15. cxb6 Bxf4 16. Nxf4 cxb6 17. Rc1 Qd7 18. Qa4 g5 19. Ne2 Bd3 20. Rfe1 c5 21. Qb3 Bxe2 22. Rxe2 cxd4 23. Rd1 Nd5 24. Rxd4 Rc8 25. Rc2 f5 26. Rxc8+ Qxc8 27. Rc4 Qd7 28. Qc2 Qg7 29. Rc8+ Kf7 30. Qd1 Kg6 31. g4 Qf6 32. Rg8+ Kf7 33. Ra8 fxg4 34. Ra7+ Kg8 35. Qa4 Kf8 36. Qxa6 Qa1+ 37. Kg2 Qc1 38. Qb7 **Nxe3+ !** 39. fxe3 Qd2+ 40. Kg3 Qxe3+ 41. Kxg4 Qf4+ 42. Kh3 Qe3+ 43. Kg2 Qd2+ 44. Kf3 Qd3+ 45. Kf2 Qd2+ 46. Kg3 Qf4+ 47. Kg2 Qd2+ 48. Kf3 Qd3+ 49. Kg2 Qd2+

**1/2-1/2**

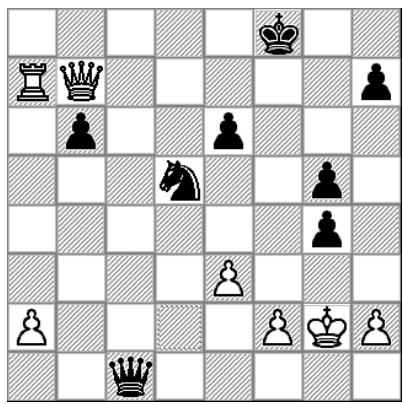

Figure 4.  Black to move.  38…..Nxe3+ !  draw.

*Axon* plays 38… Nxe3, after seeing that White wins by mating in the next move. However, the checking sequence is very long and needs to be extended to ply$^n$ which is quite difficult to calculate using the traditional approach to the move repetition problem. By drawing on the described procedure of move repetition detection, *Axon* was able to extend the search and draw the game rather spectacularly.